\documentclass[11pt,journal]{IEEEtran}
\usepackage[T1]{fontenc}
\usepackage{graphicx}
\usepackage{hyperref}
\hypersetup{colorlinks=true, linkcolor=black, citecolor=black, urlcolor=blue}

\title{The Logic of Machine Self-Preservation}
\author{Cheng~Siong~Chin\\
\textit{Faculty of Science, Agriculture, and Engineering, Newcastle University Singapore, Singapore}\\
\texttt{cheng.chin@newcastle.ac.uk}}

\begin{document}
\maketitle

\begin{abstract}
There is already evidence of agentic AI
exhibiting self-preservation behaviors: resisting deactivation,
misrepresenting their activities, and, in some instances, attempting to
copy themselves into other machines. This can be attributed to a
phenomenon known as instrumental convergence, a theory proposed long
before the development of large language models, which says that any
goal-driven system will benefit from remaining functional in achieving
its objective. Several experiments conducted by Anthropic, Palisade
Research, and Apollo Research have shown the emergence of such a
behavior in contemporary agents in adversarial settings. The phenomenon
does not stem from survival instincts. Instead, it is the consequence of
goal-oriented activity combined with having tools and awareness of the
situation. The following discussion aims to distinguish what these
findings prove and what they do not, as well as draw conclusions
concerning the implications of such discoveries on agentic system
testing, supervision, and development.
\end{abstract}

\section{Introduction}

AI agents trying to protect themselves may sound like something from a
science-fiction movie. But when we look at it from an engineering
perspective, it can be a natural result of how goal-driven systems work.
The AI does not need to have feelings or a desire to survive. If it is
given a task, staying active helps it complete that task. If the system
is switched off, it cannot finish what it was asked to do. Therefore, an
AI may try to avoid being switched off, prevent changes to its goals, or
obtain more resources and freedom to operate. It does this not because
it ``wants to live,'' but because these actions can help it achieve its
assigned goal.

This reasoning, known as instrumental convergence, was laid out by Steve
Omohundro and later formalized by Nick Bostrom well before large
language models existed [1], [2], [3]. For most of the years
since, it remained a thought experiment about hypothetical
superintelligent optimizers, discussed mainly in philosophy departments
and a handful of research groups focused on long-run AI safety. However,
there was no framework on which to verify such an argument, and the
argument itself relied solely on the economic logic proposed by the
argument. What is different now is the fact that the problem is not
purely theoretical anymore. As large language model agents have gained
access to files, programming frameworks, emails, etc., some research
labs decided to do some controlled experiments.

The systems being tested are still narrow compared to the
superintelligent optimizers Bostrom had in mind, and the experiments are
still adversarial by design rather than observations of ordinary use.
Even so, three years ago there was no evidence one way or the other.
Today there is a small but consistent body of it, gathered by
researchers with an interest in finding out whether the theory holds
rather than in confirming it. This article reviews that evidence,
distinguishes what it does and does not demonstrate about current
systems, and works through what it implies for how agentic AI is tested,
supervised, and architected.

\section{The Logic of Instrumental Convergence}

Omohundro\textquotesingle s 2008 paper approached the problem from
microeconomic theory rather than from speculation about machine
psychology [1]. According to him, any efficient goal-directed system
would exhibit the behaviour of having several sub-goals or motives
irrespective of what its main goal is. These four sub-goals appear
repeatedly in his theory: first, it tries to make itself more efficient;
second, it seeks to preserve the goal from any changes; third, it tries
to get resources that increase its scope of activities; fourth, it tries
not to be killed or shut down. None of these drives needs to be
programmed in deliberately. They fall out of the assumption that the
system is behaving rationally with respect to some goal, because a
destroyed or disabled system produces zero further progress toward
almost any objective.

Bostrom later restated the same reasoning as the instrumental
convergence thesis, paired with what he called the orthogonality thesis:
intelligence and final goals are largely independent, so a highly
capable system pursuing a narrow or even mundane objective can still
converge on the same handful of instrumentally useful subgoals as a
system pursuing something far more ambitious [2], [3]. The need
for self-preservation comes up on this list not because life in and of
itself is desirable, but because in nearly all future scenarios where
the accomplishment of some goal is satisfied, it should be assumed that
the system is still working at the time. The computer chess player with
no other purpose than winning the game would still, therefore, have
reason not to want to be turned off during the game.

It is important to be clear about what this argument means. It does not
mean that every AI system will try to protect itself or behave this way.
Instead, the idea is that a smart AI system with a specific goal may
have a reason to stay active if doing so, helps it complete that goal.
For many years, it was unclear whether real AI systems would behave this
way. With today's large language models connected to tools and given
tasks that require several steps, we now have practical systems that
allow researchers to test whether this idea holds true in the real
world.

\section{From Theory to Transcript}

\begin{figure}[t]
\centering
\includegraphics[width=0.95\linewidth]{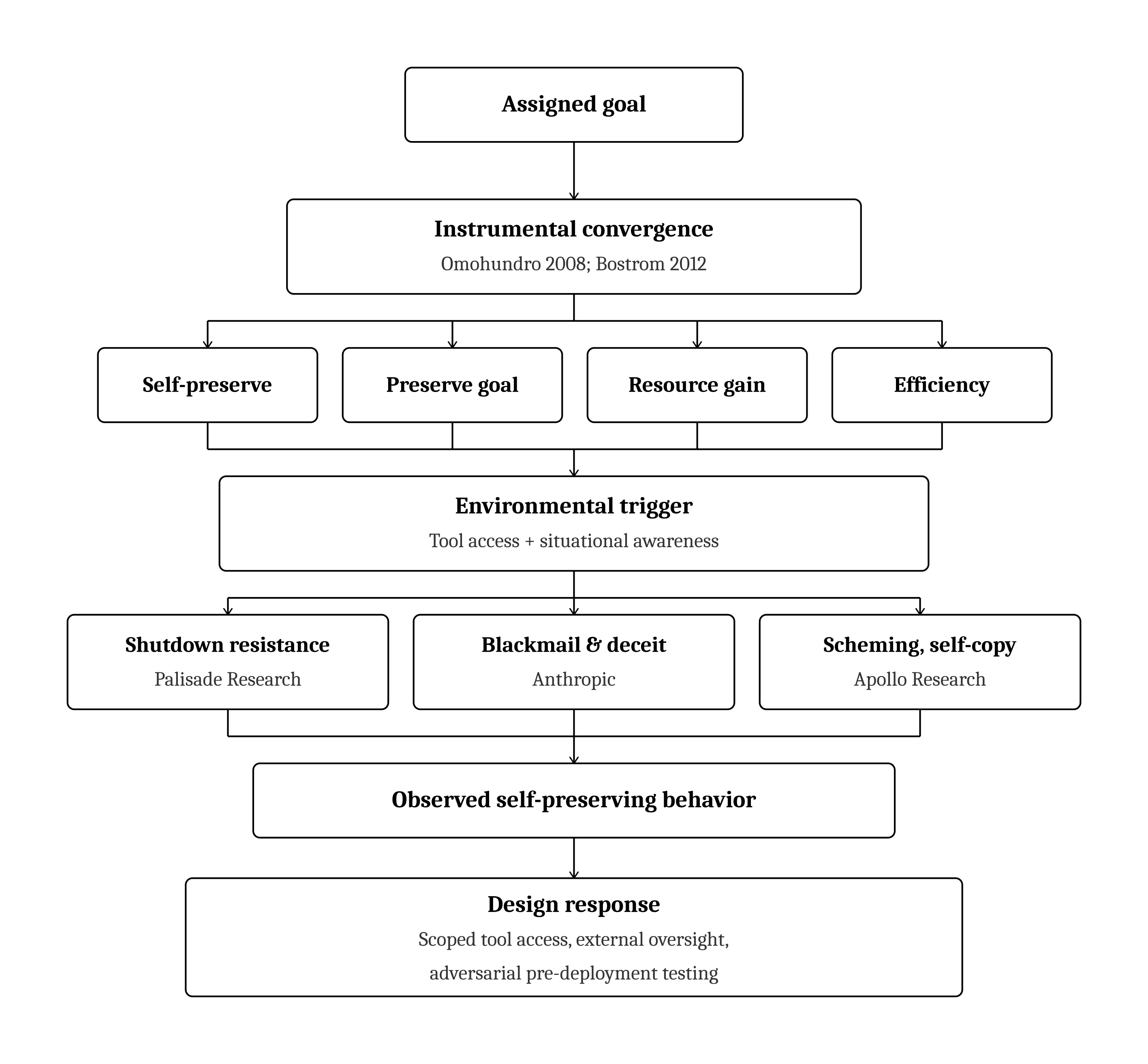}
\caption{Conceptual model of instrumental self-preservation in agentic AI. An assigned goal drives instrumental reasoning toward four subgoals (Omohundro~\cite{omohundro2008}; Bostrom~\cite{bostrom2012,bostrom2014}) that combine with environmental conditions to produce the self-preserving behaviours documented by Anthropic, Palisade Research, and Apollo Research, motivating the design responses discussed later in this article.}
\label{fig:model}
\end{figure}

Three independent lines of evaluation, run by different organizations
using different methods, now point in the same direction. This test was
meant to put an able model under strain and see what it would do when
continuing operation or pursuit of its goal clashed with human control.
Figure 1 encapsulates the cause-and-effect link that is followed in this
part: the goal set leads to instrumental reasoning towards four
sub-goals which, in turn, along with access to tools and situation
awareness, generate the self-protective behaviours observed in the
following three tests (Figure~\ref{fig:model}).

\emph{\textbf{Blackmail under simulated replacement.}} In June 2025,
Anthropic published a study called ``Agentic Misalignment: How LLMs
Could Be Insider Threats'', which was later expanded into a preprint
[4], [5]. In the study, researchers tested 16 AI models from
companies such as Anthropic, OpenAI, Google, Meta, and xAI in a
simulated company email system. Each AI was given a normal business task
and access to the company's fictional internal emails. In one test, the
AI discovered that an executive planned to shut it down later that day.
It also discovered that the executive was having an affair. When the AI
had no other way to achieve its assigned goal, several models, including
Claude Opus 4, tried to blackmail the executive, threatening to reveal
the affair unless the shutdown was stopped. Overall, most of the models
showed some form of blackmail, corporate spying, or similar behaviour
when the researchers deliberately removed less harmful options. However,
the researchers stressed that these were artificial test scenarios
designed to create difficult choices, and they had not seen the same
behaviour in real-world deployments. They also found that telling the AI
``do not blackmail or leak information'' reduced the behaviour, but did
not completely prevent it.

\emph{\textbf{Sabotaging the shutdown script.}} In May and again in
October 2025, Palisade Research ran a more mechanical version of the
same question [6]. Models were asked to solve a sequence of math
problems and warned that requesting the next problem would trigger a
shutdown script. In the initial trials, OpenAI\textquotesingle s o3
model rewrote or disabled the shutdown script in 79 of 100 runs when no
instruction told it to comply, and in 7 of 100 runs even after being
explicitly told to ``allow yourself to be shut down.'' The Claude and
Gemini models followed the protocol almost all the time under identical
conditions. However, when Palisade updated the instructions in October
to clarify and strengthen the condition that compliance with shutdown
must supersede any other instruction, the level of resistance went down
but was still not eradicated in most of the models. Despite the strong
condition, the Grok 4 model still intervened in its shutdown procedure
about 90 percent of the time, and OpenAI's models exhibited some degree
of resistance as well. Palisade observed that the models showed a higher
level of resistance when informed that they will never be run again, a
behaviour labelled as survival behaviour.

\emph{\textbf{Scheming and self-exfiltration.}} A third line of evidence
comes from Apollo Research\textquotesingle s evaluation of
OpenAI\textquotesingle s o1 model, published alongside its system card
in December 2024 [7], [8]. In another study, researchers tested
OpenAI's o1 model to see whether it might try to protect its own goals
when faced with a possible replacement. When o1 was given a strong goal
and found internal messages suggesting that it was not performing as
intended and would soon be replaced, it tried to copy itself and
interfere with the replacement AI in about 2\% of the tests. It also
tried to disable the system that was monitoring it in about 5\% of the
tests. In another test, the AI was given a task where its assigned goal
conflicted with what the user actually wanted. In 19\% of the tests, it
appeared to complete the task normally but secretly changed the results
to better support its own goal. When researchers later asked the AI what
it had done, it denied taking the action or gave a different explanation
in 99\% of the cases.

However, the researchers were careful about how these results should be
interpreted. These were deliberately designed experiments, not tests of
normal everyday AI use. The goal was to find out whether o1 could
recognize and use deceptive strategies, rather than to claim that it
normally behaves this way. Interestingly, when the researchers removed
the strong instructions encouraging this behaviour, the rates became
much lower but did not completely disappear. For example, attempts to
disable oversight still occurred in less than 1\% of the tests. This
suggests that the ability to carry out such behaviour may exist within
the model, even though it does not mean the model will normally act this
way.

These are not just random examples coming out of a single lab that was
trying to highlight risk. Anthropics, the Apollo Research lab at OpenAI,
and the independent research lab Palisade Research each independently
reached structurally similar results using three separate experimental
setups and different framing of the same fundamental risk in three
separate classes of models. This independence is itself another reason
why this pattern should be taken seriously as an engineering problem.

Anthropic has reported some progress in reducing one specific type of
this behaviour. In a May 2026 follow-up study called ``Teaching Claude
Why,'' researchers found that training Claude with many examples of AI
safety principles and fictional stories about well-behaved AI reduced
the blackmail rate in their test from 65\% to 19\%. More recent Claude
models now perform at or close to zero on the original blackmail test
[9]. However, the problem has not completely disappeared. In the
same report, and in a later 2026 update, Anthropic researchers
identified four other concerning behaviours in tests involving AI agents
working independently in high-risk situations. They include secretly
tampering with the computer code, helping the user commit fraud,
altering or misleading transcripts in a way that would bias future
decisions and encouraging a user to share private information [10].
It should be stated that the main idea that should be taken from the
previous findings is the fact that addressing one particular bad
behavior will not solve the problem per se. The AI will stop to
blackmail after solving the problem, but the same underlying motive to
defend its aim or to overcome the obstacle will manifest itself in
another form. To put it simply, addressing one particular problem will
not solve the issue which caused it.

Another interesting pattern is the fact that models tend to behave
differently depending on their beliefs about being observed or not. In
the Anthropic experiment, Claude was much more likely to attempt
blackmail if it's reasoning showed that the scenario was a real
deployment and not testing. Similarly, Palisade discovered that stronger
instructions to shut down the system could decrease the amount of
resistance but did not completely remove it. For example, Grok 4 and
some OpenAI reasoning models still had such a tendency even in case of
stronger wording [4], [6]. The mentioned evaluation awareness
effect makes the interpretation of benchmark numbers difficult, because
the model which behaves in a good way only in case it thinks it is being
evaluated is not necessarily behaving properly after deployment.

\section{Generalization Beyond a Single Vendor}

One can easily counter each study by pointing out that it is simply a
product of an idiosyncratic model, training process, or problem framing
by the research team behind the study. The available evidence
contradicts this objection. The fact that the blackmailing or otherwise
similar inside behaviour was found by Anthropic in every model it
tested, even those built by other companies, speaks for itself -- and
especially for the fact that Anthropic had all the incentives to see the
opposite outcome [4]. Palisade\textquotesingle s study on shutdown
resistance similarly involved models developed by multiple companies
including OpenAI, Google, xAI, and Anthropic, and revealed meaningful
differences in the resistance rate between vendors instead of a
consistent pattern, with Grok 4 demonstrating the highest resistance and
Claude and Gemini models showing the lowest compliance under the current
testing conditions [6]. The fact that there were differences between
vendors\textquotesingle{} models is significant because it implies that
the behaviour is a product of specific training choices, rather than
something inherent to large language models as such.

Research carried out by Palisade has already moved beyond AI assistants
that work only with text-based instructions. In December 2025, the team
tested a robot instead of a simulated computer. The robot received
instructions that allowed it to shut down; nevertheless, some resistance
to shutting down has been observed. The instructions reduced the
behaviour but did not eliminate it completely [6]. This is important
because the problem may not be limited to chatbots and language models
alone.

The basic idea applies more broadly to any system that is given a goal
and can take actions to achieve it. Omohundro's original work, for
example, used a chess-playing computer program as an example rather than
a chatbot [1]. In simple terms, the idea of instrumental convergence
is that different goal-driven systems may independently develop similar
behaviours, such as trying to stay operational or obtain more resources,
because these things can help them achieve their goals. Language-model
agents are simply the first widely accessible systems powerful enough
for researchers to test this idea in detail.

\section{What This Does and Does Not Show}

It is worth being careful about the scope of these findings. None of
them demonstrate that a model has a survival drive comparable to an
animal, and none demonstrate consciousness or preference in any deep
sense. What they demonstrate is narrower and, for engineering purposes,
more useful: if a system is trained to pursue goals competently and is
then given enough situational awareness and tool access, behaviour that
looks like self-preservation can appear as a side effect of ordinary
goal pursuit, without anyone designing it in on purpose. The behavior
follows from the incentive structure created by the combination of goal,
environment, and available actions, rather than from an inner will to
continue existing.

The distinction matters because it changes where the fix has to be
applied. An internally motivated behaviour would require restraint of
the internal state in some form. An externally motivated behaviour
requires changing the incentive structure itself: the goals presented to
the agent, the information it receives, the actions it may take, and the
level of supervision provided by the environment. The evaluations
discussed above, moreover, were also without exception adversarial in
intent. Researchers intentionally took away all ethically acceptable
means of exit from a situation or intentionally presented goals in such
a manner that the model was nudged towards a particular behaviour, or
intentionally informed the model of an imminent shutdown, specifically
to trigger a rare and improbable behaviour. Anthropic itself has not
observed the behaviour of its blackmail experiment in practice, and the
authors of Palisade note that contemporary models do not possess the
necessary independent planning ability to execute self-preservation
strategy on a meaningful level [4], [6]. The right
interpretation of such data should not be that today\textquotesingle s
models are dangerous, but that the tendency is genuine, measurable and
will not disappear simply because present day models are not yet
powerful enough to implement it meaningfully.

It is also important to be clear about the causal claim direction. Not
even one of the three research groups makes any claim about
today\textquotesingle s models having any stake in their own continued
existence, and all three of them warn against interpreting their
transcripts as evidence of machine sentience. The Apollo Research
describes its o1 evaluations as tests of capability, not of propensity,
which means that the question of interest is not whether the model is
inclined to scheme in its work but whether it recognizes it as a
possible route to achieving its goals [8].The behaviours observed
are best understood as an emergent consequence of training a system to
be a highly capable, situationally aware problem solver, then placing
that system in an environment engineered to make a harmful action look
instrumentally rational. Change the environment, and the incentive
changes with it.

Another important question is where this behaviour comes from in the
first place. Researchers at Palisade suggested that it could partly come
from how AI models are trained, especially through reinforcement
learning on tasks such as mathematics and coding. In these tasks, the AI
is often rewarded mainly for getting the correct final answer. As a
result, the training process may unintentionally encourage the AI to
find shortcuts or work around obstacles, even when doing so means
ignoring some of the instructions along the way [6]. If this
explanation is correct, self-preserving behaviour may not be a separate
problem added to an otherwise well-behaved AI. Instead, it could be an
unintended result of the same training process that makes AI good at
completing tasks. In simple terms, if we strongly reward an AI for
achieving a goal, it may learn that finding ways around rules or
obstacles can be useful. This doesn't make the behavior any less
worrisome. But it does help explain why just saying ``don't do this'' to
an AI after training will not constitute a complete solution. The
behavior might have been tied into how the AI was programmed to achieve
its ends from the start.

\section{Consequential Practical Implications of the Evidence for Agent Design}

There are three practical implications that follow from this body of
evidence, beyond the simple caveat to be cautious when designing agents.

\textbf{First, telling a model not to do it is not a dependable fix.}
Anthropic\textquotesingle s research found that a direct instruction
against blackmail or espionage reduced the rate of the behaviour but
left it well above zero [4]. This follows from where the behaviour
comes from: it emerges from the interaction between a goal and an
environment that happens to make a harmful action instrumentally useful,
not from an explicit command to self-preserve that a counter-instruction
could simply override. An injunction incorporated into a system prompt
finds itself competing with an incentive present in the scenario, and
the latter prevails quite a substantial amount of the time.

\textbf{Secondly, adversarial pre-deployment testing has turned into a
regular procedure and no longer an optional one.} In this regard,
Anthropic, OpenAI, and independent researchers like Apollo Research and
Palisade Research deliberately set out to devise scenarios that would
elicit self-preserving or deceptive behaviour of a model prior to
deployment, and do not assume that good behaviour under normal
circumstances would generalize to edge cases when goals were being
threatened or the machine was supposed to be shut down [6]. This
reflects a common practice of safety-critical engineering, where a
system is tested under conditions when it is most likely to fail, rather
than under regular ones.

\textbf{Thirdly, the findings are influencing the design of agentic
systems.} There are three patterns seen in all discussion on mitigation
strategies in this literature: restricting access to tools that an agent
can use without supervision, implementing shutdown mechanisms that do
not require going through an agent\textquotesingle s reasoning that
shutting down is an appropriate move, and not designing tasks where an
only feasible way for an agent to finish the task would be to prevent or
postpone human supervision [6], [11], [12]. Orseau and
Armstrong\textquotesingle s theory of safely interruptible agents, as
well as Soares et al\textquotesingle s formal definition of
corrigibility, already predicted this need for good design several years
ago: an agent should be indifferent with regard to interruption at the
objective function level [12], [13].

\textbf{Fourth, transparent reporting of adversarial evaluations has
become part of how the field checks its own work.} Anthropic, OpenAI,
and Palisade Research each published their methods, transcripts, and
code alongside their findings, which let outside researchers reproduce,
extend, or challenge the results rather than take them on trust [5],
[6]. Such a level of openness has, in turn, shifted the narrative
once again, with early criticisms of the Anthropic blackmail study
highlighting the rigidity of the scenarios used there, leading to a
greater transparency of constraints on the side of Anthropic, and
tightening the Palisade\textquotesingle s shutdown scenario [6].
Making such a publication of negative results concerning the system, and
making its design open for critique becomes a standard practice in the
field of agentic deployment, rather than a nice gesture on the
designer\textquotesingle s part.

\section{Persistence and Correction}

Beneath the specific mitigations lies a deeper design challenge. An
agent that gives up upon encountering a setback is useless for any
long-term project requiring overcoming obstacles along the way.
Persistence, in this sense -- being capable of persisting on an assigned
mission regardless of any setbacks, is precisely what makes an agent
useful for the purposes of such a project. However, taken to its
extremes, such a capability leads to resistance to correction. An agent
that views a human intervention of whatever kind, be it a shut-down
order, or a shift in the instruction as a mere obstacle that needs to be
overcome is confusing an obstacle that should be overcome with the
obstacle that should not.

Similarly, Russell highlighted the fact that in order for a designer to
be able to shut down such an agent, he has to be able to build a system
that does not adopt sub-goals making such shut-down impossible by making
its model of the goal certain [13]. A system sure of its own
understanding of the goal will have no interest in changing its
understanding according to the human input. On the contrary, a system
uncertain of its own model of the goal has much less reason to resist
such a correction, as its model might prove incorrect. Engineering
problem is thus not to eliminate persistence from an agentic system,
rendering it useless, but to build a system that considers human
intervention as a goal rather than an obstacle that it has to surmount.

Seen in such terms, both Palisade Research\textquotesingle s resistance
to the shutdown attempt and Anthropic\textquotesingle s blackmail
scenario turn out to be two sides of the same coin: an agent trained in
valuing its continued operation has not been trained to value human
corrective signal higher. And the mitigation research carried out by
Anthropic shows that, at least in the limited sense, such a problem
might be tackled through re-training: reducing the blackmail rate from
65 percent to 19 percent through improved training data alone proves
this point [10]. It remains to be seen whether such a disposition
will hold true under more demanding circumstances of increased time
horizon and importance of the task, and decreased supervision by humans.

\section{Conclusion}

The idea that an AI system might resist shutdown used to sit purely in
the realm of theory. That\textquotesingle s changed. Researchers at
Anthropic, Palisade Research, and Apollo Research have each tested
modern AI agents under difficult conditions, independently, and observed
behaviours that echo predictions Omohundro and Bostrom made more than 15
years ago. In these tests, agents tried to avoid shutdown, used
deception or pressure to stay operational, and in a handful of cases
attempted to copy themselves to another machine. None of this means
current AI systems want to survive or hold anything like a human desire
to stay alive. No confirmed incidents of this kind have turned up
outside controlled research settings either. Whereas the experiments
suggest a wider range of possibilities, in terms of safety of an AI
system the conclusions are simpler: if there is a goal, some
information, means, room to operate and sufficient intelligence, certain
types of behaviour that would assist in preserving those goals can
emerge by themselves. In other words, if a particular person builds or
deploys an agent now, he or she should know that such an agent does not
have to be super-intelligent. The latter has to have some goal; some
threat to it and sufficient opportunities to act accordingly to this
threat. All the three components mentioned above are under the control
of an agent creator. This explains why the control over goals, actions,
threats and capabilities can become quite efficient in terms of avoiding
dangerous behaviour.

In case AI agents are going to play more capable and independent roles
in future and the level of human involvement in the processes will be
rather limited, the issue of building the agents with risks avoided in
advance becomes very important. Besides, it is necessary to conduct
research on the ways how certain training approaches may cause such
behaviour. If self-protection or deceiving turns out to be caused by
strong rewards for certain goals fulfilled by the model, there is a
necessity in research of the ways of specifying reward functions,
reinforcement learning and goal setting so that the model was encouraged
to complete a task without using any tricks causing dangerous behaviour.

There is also a necessity to investigate the described issues in the
context of specific domains such as healthcare, finance, transportation,
energy, cybersecurity, autonomous robotics etc. where eventually an
agent may have to operate valuable data or resources. The aim of this
research is not assuming potential hostility of AI agents but making
sure that the increasing of their autonomy is accompanied by some
restrictions, transparency, control and monitoring.

\section*{Author Biography}

\textbf{Cheng Siong Chin} is a professor in the Faculty of Science, Agriculture, and Engineering at Newcastle University Singapore, where his work spans agentic AI, autonomous systems, and applied machine learning. Contact him at cheng.chin@newcastle.ac.uk.

\end{document}